\documentclass[runningheads]{llncs}
\usepackage[T1]{fontenc}
\usepackage{CJK}
\usepackage{graphicx}
\begin{document}
\title{Topic Shift Detection in Chinese Dialogues: Corpus and Benchmark}
%
%

\author{Jiangyi Lin\inst{1} \and
Yaxin Fan\inst{1} \and
Feng Jiang\inst{2} \and
Xiaomin Chu\inst{1} \and
Peifeng Li\thanks{Corresponding author.}\inst{1}}
\authorrunning{Lin. et al.}
%
\institute{School of Computer Science and Technology, Soochow University, Suzhou, China
\email{\{jylin, yxfansuda\}@stu.suda.edu.cn}, 
\email{\{xmchu, pfli\}@suda.edu.cn}\\
\and School of Data Science, The Chinese University of Hong Kong, Shenzhen, China\\
\email{jeffreyjiang@cuhk.edu.cn}}

\maketitle              
\begin{abstract}
Dialogue topic shift detection is to detect whether an ongoing topic has shifted or should shift in a dialogue, which can be divided into two categories, i.e., response-known task and response-unknown task. Currently, only a few investigated the latter, because it is still a challenge to predict the topic shift without the response information. In this paper, we first annotate a Chinese Natural Topic Dialogue (CNTD) corpus consisting of 1308 dialogues to fill the gap in the Chinese natural conversation topic corpus. And then we focus on the response-unknown task and propose a teacher-student framework based on hierarchical contrastive learning to predict the topic shift without the response. Specifically, the response at high-level teacher-student is introduced to build the contrastive learning between the response and the context, while the label contrastive learning is constructed at low-level student. The experimental results on our Chinese CNTD and English TIAGE show the effectiveness of our proposed model.

\keywords{Dialogue topic shift detection  \and Hierarchical contrastive learning \and Chinese natural topic dialogues corpus.}
\end{abstract}

\section{Introduction}
Dialogue topic shift detection is to detect whether a dialogue's utterance has shifted in the topic, which can help the dialog system to change the topic and guide the dialogue actively. Although dialog topic shift detection is a new task, it has become a hotspot due to its remarkable benefit to many downstream tasks, such as response generation \cite{dai2021dialogue} and reading comprehension \cite{li2021dadgraph,li2021self}, and can help those real-time applications produce on-topic or topic-shift responses which perform well in dialogue scenarios \cite{ghandeharioun2019approximating,einolghozati2019improving,liu2018dialogue}.

\begin{figure}[htbp]
\vspace{-0.4cm}
\centerline{\includegraphics[scale=0.32]{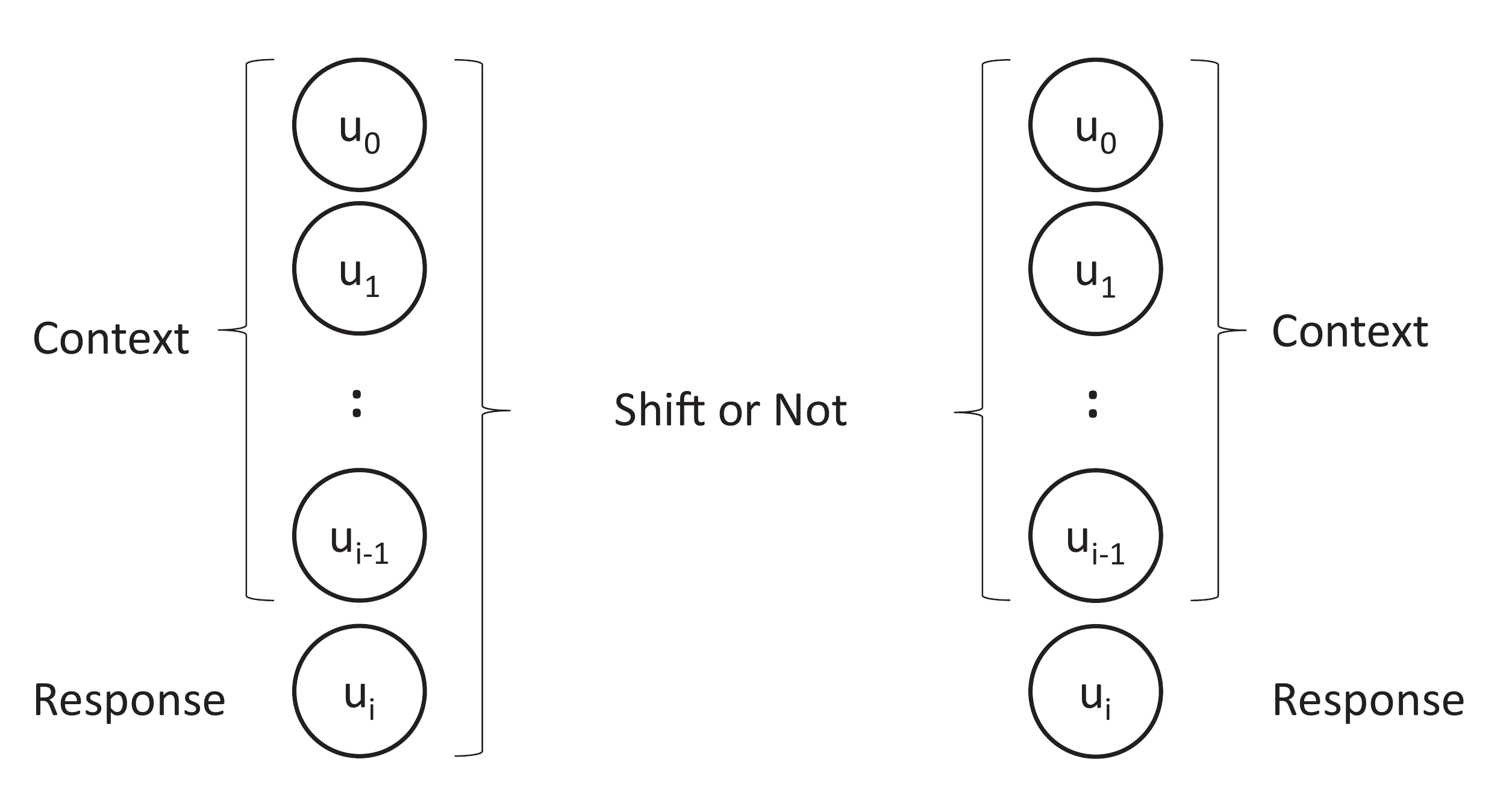}}
\caption{Two lines of dialogue topic shift detection tasks to detect whether it exists topic shift between the utterances $u_{i-1}$ and $u_i$, where the response-known task (left) can use the response $u_i$, while the response-unknown task (right) can be regarded as topic shift prediction without the response $u_i$.}
\label{fig1}
\vspace{-0.3cm}
\end{figure}

The task of dialogue topic shift detection can be divided into two lines, i.e., response-known task and response-unknown task, as shown in Fig.~\ref{fig1}. The former can gain the response information and obtain a better result, while the latter is the opposite. 
Moreover, both of them are not accessible to future information. This is the biggest difference from the task of text topic segmentation, in which all the basic utterances are visible to each other. That is, those existing topic segmentation models cannot be applied to dialogue topic shift detection since it depends on the response and its subsequent utterances heavily. Therefore, it is more difficult to discern differences between utterances in the task of dialogue topic shift detection. Due to the absence of future utterances, dialogue topic shift detection is still a challenging task. 

In this paper, we focus on the response-unknown task of topic shift detection in Chinese dialogues. There are two issues in the response-unknown task of topic shift detection in Chinese dialogues, i.e., lack of annotated corpus in Chinese and how to predict the response. 

There are only a few publicly dialogue topic shift corpus available and most of them are provided for the segmentation task, which does not satisfy natural
conversation. Xie et al. \cite{xie2021tiage} provided a detailed definition of the dialogue topic shift detection task, and annotated an English dialogue topics corpus TIAGE. Although it can fill the gap in the corpus of English conversation topics, its scale is still too small. In Chinese, Xu et al. \cite{xu2021topic} annotated a Chinese dialogue topic corpus.
However, due to its small size and poor quality, this is detrimental to the further research and development of Chinese dialogue topic shift tasks. To fill the gap in the Chinese natural dialogue topic corpus, we first annotated a Chinese Natural Topic Dialogue (CNTD) corpus which consists of 1308 dialogues with high quality. 

Xie et al. \cite{xie2021tiage} also established a benchmark for this response-unknown task based on the T5 model \cite{raffel2020exploring} and this benchmark only used the context to predict topic shift and performed poorly due to the lack of the response information.
Thus, it is more challenging to predict the topic shift in natural dialogue without useful response information. 


The teacher-student framework has been used widely to obtain information that is not available to the model \cite{dai2021dialogue}. To solve the issue of the lack of response information, we propose a teacher-student framework to introduce the response information. The teacher can obtain the response information, and the student can learn the response information from the teacher through knowledge distillation. To facilitate knowledge transfer, the student mimics the teacher on every layer instead of just the top layer, which alleviates the delayed supervised signal problem using hierarchical semantic information in the teacher \cite{li2019hint}.

Besides, we construct hierarchical contrastive learning in which we consider the teacher-student as high-level and the student as low-level. At high-level, we build an information simulation loss between the context and the response to improve the semantic information of the student model with more reliable predictive information. At low-level, we design a semantic coherence-aware loss to better distinguish the different shift cases and produce more reliable prediction results.

Finally, the experimental results on our Chinese CNTD and the English TIAGE show that our proposed model outperforms the baselines. 
The contributions of this paper are as follows.
\begin{itemize}
\vspace{-0.1cm}
\item We manually annotate a corpus with 1308 dialogues based on NaturalConv to fill the gaps in the Chinese natural dialogues topic corpus.
\item We propose a teacher-student framework to learn the response information for the topic shift detection task.
\item We introduce hierarchical contrastive learning to further improve performance.
\item The experimental results both on the CNTD and TIAGE datasets show that our model outperforms the baselines.
\end{itemize}

\section{Related Work}

\subsection{Corpus}
Previous studies explored the dialogue topic tasks and published the annotated topic dialogue corpus. For English, Xie et al. \cite{xie2021tiage} annotated the TIAGE consisting of 500 dialogues with 7861 turns based on PersonaChat \cite{zhang2018personalizing}. Xu et al. \cite{xu2021topic} built a dataset including 711 dialogues by joining dialogues from existing multi-turn dialogue datasets: MultiWOZ Corpus \cite{budzianowski2018multiwoz}, and Stanford Dialog Dataset \cite{eric2017key}. Both corpora are either small or limited to a particular domain, and neither applies to the study of the natural dialogue domain. 

For Chinese, Xu et al. \cite{xu2021topic} annotated a dataset including 505 phone records of customer service on banking consultation. However, this corpus is likewise restricted to a few specialized domains while natural dialogues are more complicated. Natural dialogues have a range of topic shift scenarios, unrestricted topics, and more free colloquialisms in the utterances. The above corpus is insufficient to fill the gap in the Chinese natural dialogue topic corpus.

\subsection{Topic Shift Detection in Dialogues}
The task of dialogue topic shift detection is also in its initial stage and only a few studies focused on this task. As we mentioned in Introduction, topic segmentation is a similar task. Hence, we first introduce the related work of topic segmentation in dialogues.
Due to the lack of training data on dialogue, early approaches of dialogue topic segmentation usually adopted an unsupervised approach using word co-occurrence statistics \cite{eisenstein2008bayesian} or sentence topic distributions \cite{du2013topic} to measure sentence similarity between turns to achieve detection of thematic or semantic changes. Recently, with the availability of large-scale corpora sampled from Wikipedia, supervised methods for monologic topic segmentation have grown rapidly by using partial tokens as ground-truth segmentation boundaries, especially neural-based methods \cite{koshorek2018text,badjatiya2018attention,arnold2019sector}. These supervised solutions are favored by researchers due to their more robust performance and efficiency.

Dialogue topic shift detection is strongly different from dialogue topic segmentation. For the dialogue topic detection task, Xie et al. \cite{xie2021tiage} proposed a detailed definition with two lines: the response-known task considering both the context and the response, and the response-unknown task considering the context only. However, methods based solely on the context are still scarce. Only Xie et al. \cite{xie2021tiage} predict the topic shift or not based on the T5 model. Sun et al. \cite{sun2019topic} introduce structural and semantic information to help the model detect topic shifts in online discussions, which is similar to response-known task. It is imperative to address the dialogue topic shift detection. In general, the dialogue topic shift detection task is still a challenge, as it can only rely on the context information of the dialogue. In this paper, we solve the lack of response information by utilizing knowledge distillation and hierarchical contrastive learning.

\section{Corpus}
The existing corpus of Chinese dialogue topic detection \cite{xu2021topic} is small and does not satisfy natural conversation. Although the English dialogue topic corpora can be converted into Chinese by machine translation, they lack natural conversation colloquiality and are small in size. Therefore, we annotate a Chinese dialogue topic detection corpus CNTD based on NaturalConv dataset \cite{wang2021naturalconv}.

In this section, we show our annotation guidelines and outline the reasons for our selection of corpus sources, as well as the manual annotation procedure and data statistics. We also analyze the topic shift distribution in CNTD.

\subsection{Strengths}Each dialogue in our corpus has a piece of news as a base document, which is not available in other corpus and can be used as additional information for further research and expansion. The news is from six domains, which brings our conversations closer to natural dialogue. Besides, the speakers in our corpus are not restricted in any way, which also makes it closer to natural dialogues. In addition, we annotated the fine-grained dialogues topics, refer to Section 3.2. Fine-grained labels are beneficial to promote further research on dialogue topics.

Compared with the existing Chinese topic corpus annotated by Xu et al.  \cite{xu2021topic}, the dialogues in our corpus do not have meaningless and repetitive turns. Also, the corpus is more than twice the size of the other corpus. In addition, the news in the corpus can be studied as additional information for the dialogues.

\subsection{Annotation Guidelines}

Following the annotation guidelines in TIAGE\cite{xie2021tiage}, we distinguish each dialogue turns whether changed the topic compared with the context. The response of a speaker to the dialogue context usually falls into one of the following cases in dialogues where the examples can be found in Table~\ref{fig2}. 
\renewcommand{\dblfloatpagefraction}{.9}
\setlength\tabcolsep{18pt}
\begin{table}
\vspace{-0.5cm}
\caption{Different scenarios of response in dialogues.}
\begin{center}
\vspace{-0.2cm}
\begin{tabular}{lp{8cm}}
\hline
B & "Recently playing handheld games, "dunking master" to look back on the period of watching anime."\\
A & "I see this game so many platforms are pushing ah, but I have not played."\\
B & "Yeah, it’s pretty fun, you can go play it when the time comes. There are anime episodes."→ not a topic shift\\
\hline
\end{tabular}
\centerline{}
\centerline{(a) Commenting on the previous context.}
\centerline{}
\begin{tabular}{lp{8cm}}
\hline
A & "What grade is your child in?"\\
B & "He’s a freshman." → not a topic shift\\
\hline
\end{tabular}
\centerline{}
\centerline{(b) Question Answering.}
\centerline{}
\begin{tabular}{lp{8cm}}
\hline
A & "The Laval Cup is about to start, and the European team has two kings, Federer and Nadal."\\
B & "I know Federer, he is one of the best in the tennis world."→ not a topic shift\\
\hline
\end{tabular}
\centerline{}
\centerline{(c) Developing The Conversation to Sub-topics}
\centerline{}
\begin{tabular}{lp{8cm}}
\hline
B & "Haha, so what do you usually like to do sports ah? Do you usually go out for a run?"\\
A & "Rarely, usually lying at home watching TV, running and so on is to see their fat can not pretend to look."\\
B & "I also, then what TV are you watching lately? Have you been watching "Elite Lawyers"?"→ topic shift\\
\hline
\end{tabular}
\centerline{}
\centerline{(d) Introducing A Relevant But Different Topic.}
\centerline{}
\begin{tabular}{lp{8cm}}
\hline
B & "This movie I saw crying, Iron Man died, really moved."\\
A & "Yes, the special effects of this movie are very good."\\
B & "Who is your favorite actor?"→topic shift\\
\hline
\end{tabular}
\centerline{}
\centerline{(e) Completely Changing The Topic.}
\end{center}
\label{fig2}
\vspace{-1cm}
\end{table}

\begin{itemize}
\item Commenting on the previous context:
The response is a comment on what is said by the speaker previously;
\item Question answering:
The response is an answer to the question that comes from the speaker previously;
\item Developing the dialogue to sub-topics:
The response develops to a sub-topic compared to the context;
\item Introducing a relevant but different topic:
The response introduces a relevant but different topic compared to the context;
\item Completely changing the topic: The response completely changes the topic compared to the context.
\end{itemize}

Among them, we uniformly identify the two cases of greeting and farewell specific to CNTD as the topic shift.

\subsection{Data Source}
We chose the NaturalConv dataset \cite{wang2021naturalconv} as the source corpus, which contains about 400K utterances and 19.9K dialogues in multiple domains. It is designed to collect a multi-turn document grounded dialogue dataset with scenario and naturalness properties of dialogue.

We consider NaturalConv as a promising dataset for dialogue topic detection for the following reasons:
1) NaturalConv is much closer to human-like dialogue with the natural property, including a full and natural setting such as scenario assumption, free topic extension, greetings, etc.; 
2) NaturalConv contains about 400K utterances and 19.9K dialogues in multiple domains;
3) The average turn number of this corpus is 20, and longer dialogue contexts tend to exhibit a flow with more topics; 
4) The corpus has almost no restrictions or assumptions about the speakers, e.g., no explicit goal is proposed \cite{wu2019proactive}.

\subsection{Annotation Process}
We have three annotators for coarse-grained annotations and two for fine-grained annotations. Both annotations are divided into three stages as follows.

\textbf{Co-annotation Stage}
First, for coarse-grained annotations, we draw a total of 100 dialogues from each domain of the NaturalConv dataset proportionally for a total of 2014 dialogue turns. In this stage, three annotators are asked to discuss every 20 dialogues they annotated, and each annotator is asked to give a reason for the annotation during the discussion. Finally, the Kappa value of all annotators for coarse-grained annotations at this stage is 0.7426. 
In addition, we annotated the fine-grained information based on the results of the complete coarse-grained annotations. Two annotators annotated the same 150 dialogues and discussed them several times for consistency. Finally, the kappa value of all annotators for fine-grained annotations at this stage is 0.9032.
These kappa values confirm that our annotators already have sufficient annotation capabilities for independent annotation, as well as the high quality of our corpus.

\textbf{Independent-annotation Stage}
We ensured the quality of each annotator's annotation and judging criteria before starting the second phase of annotation. For both granularity annotations, we randomly assign the dialogues drawn from each domain to each annotator for independent annotation. At this stage, we annotate 1208 dialogues for coarse-grained annotations and 1158 dialogues for fine-grained annotations.

\begin{table}[h]
\caption{Category and proportion of the corpus.}
\begin{center}
\vspace{-0.4cm}
\begin{tabular}{ccccc}
\hline
\textbf{Category} & \textbf{Train}& \textbf{Val.} & \textbf{Test}& \textbf{Sum.}\\
\hline
Health(8\%) & 85 & 11 & 11 & 107\\
Education(16\%) & 167 & 22 & 21 & 210\\
Technology(17\%) & 176 & 22 & 22 & 220\\
Sports(33\%) & 347 & 45 & 46 & 438\\
Games(8\%) & 86 & 11 & 11 & 108\\
Entertainment(17\%) & 180 & 23 & 22 & 225\\
\hline
Total & 1041 & 134 & 133 & 1308\\
\hline
\end{tabular}
\label{tab2}
\end{center}
\end{table}
\begin{table}[htbp]
\vspace{-0.4cm}
\setlength\tabcolsep{27pt}
\caption{Details of CNTD.}
\begin{center}
\vspace{-0.4cm}
\begin{tabular}{cccc}
\hline
 & \textbf{Min.}& \textbf{Max.} & \textbf{Avg.}\\
\hline
Dialogue Turns & 20 & 26 & 20.1\\
Utterance Words &1 & 141 & 21.0\\
Dialogue Words & 194 & 888 & 421.7\\
Dialogue Topics& 2 & 9 & 5.2\\
Topic Turns &1 & 17 & 4.2\\
\hline
\end{tabular}
\label{tab3}
\end{center}
\vspace{-0.4cm}
\end{table}
\begin{table}[htbp]
\vspace{-0.4cm}
\setlength\tabcolsep{38pt}
\caption{Statistics for fine-grained labels.}
\begin{center}
\begin{tabular}{cc}
\hline
\textbf{Fine-grained labels} & \textbf{Count}\\
\hline
Commenting on the previous context & 15091\\
Question answering & 3505\\
Developing the dialogue to sub-topics & 857\\
Introducing a relevant but different topic & 3106\\
Completely changing the topic & 2439\\
\hline
\end{tabular}
\label{tab8}
\end{center}
\vspace{-1.0cm}
\end{table}

\textbf{Semi-automatic Rechecking Stage}
Finally, we use a semi-automatic rechecking process to ensure that the corpus is still of high quality. On the one hand, we automatically format the dialogues with annotations to detect formatting problems caused by manual annotation. On the other hand, we automatically match the related news to each dialogue and check that the topic attributes are consistent with the dialogue to rule out any possible errors.

\subsection{Annotation Results}
Due to the limited time, we randomly select 1308 dialogues from the NaturalConv dataset and annotate them with four annotators. Finally, we construct a Chinese natural topic dialogues corpus containing 26K dialogue turns.

As shown in Table~\ref{tab2}, we randomly split them into 1041 train, 134 validation, and 133 test dialogues respectively, according to the percentage of different categories. In addition, we show the details of CNTD in Table~\ref{tab3}, which shows that our corpus has enough topics and long turns which is suitable for dialogue topic detection. Finally, there are the statistics of our fine-grained labels, as shown in Table~\ref{tab8}.

\begin{figure}[htbp]
\vspace{-0.5cm}
\centering
\centerline{\includegraphics[scale=0.4]{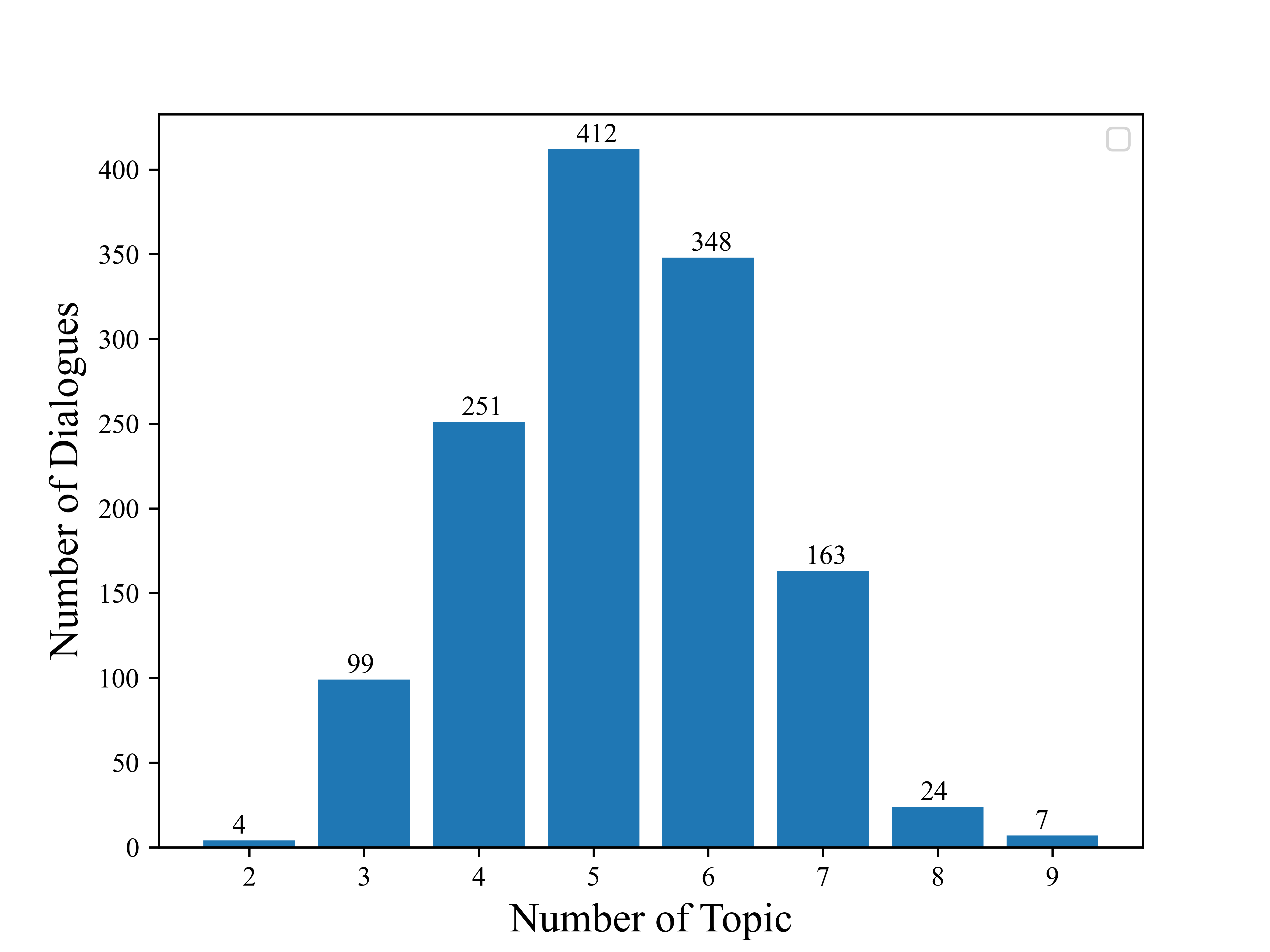}}
\caption{Number of dialogues with different numbers of topics.}
\label{fig3}
\vspace{-0.3cm}
\end{figure}
\begin{figure}[h]
\centering
\includegraphics[scale=0.4]{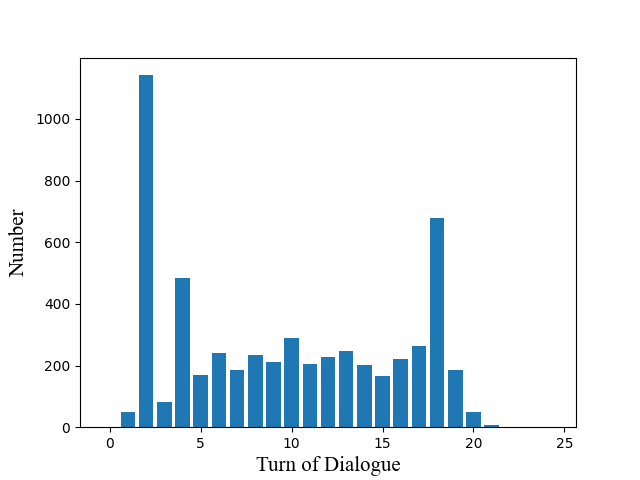}
\caption{Topic shift distribution of CNTD.}
\label{fig4}
\vspace{-0.2cm}
\end{figure}
We count the number of dialogues with different numbers of topics, as shown in Fig.~\ref{fig3}. On another side, we count the distribution of topic shift signals in dialogues, shown in Fig.~\ref{fig4}. We can see there are a total of 21 turns and three peaks of topic shift signals, which occur in $2^{nd}$, $4^{th}$, and $18^{th}$ turns, respectively. The reason is that the dialogue in our corpus usually starts with a greeting and ends with a farewell, which leads to more topic shifts at the beginning and end of the dialogues. In addition, the NaturalConv corpus gives a piece of news as the base document of the dialogue, so there are more frequent transitions from news to derived topics, leading to the third highest peak in $4^{th}$ turn. However, we think this is consistent with a natural dialogue scenario because people often talk about recent news after daily greetings.

\begin{figure*}[htbp]
\centerline{\includegraphics[scale=0.55]{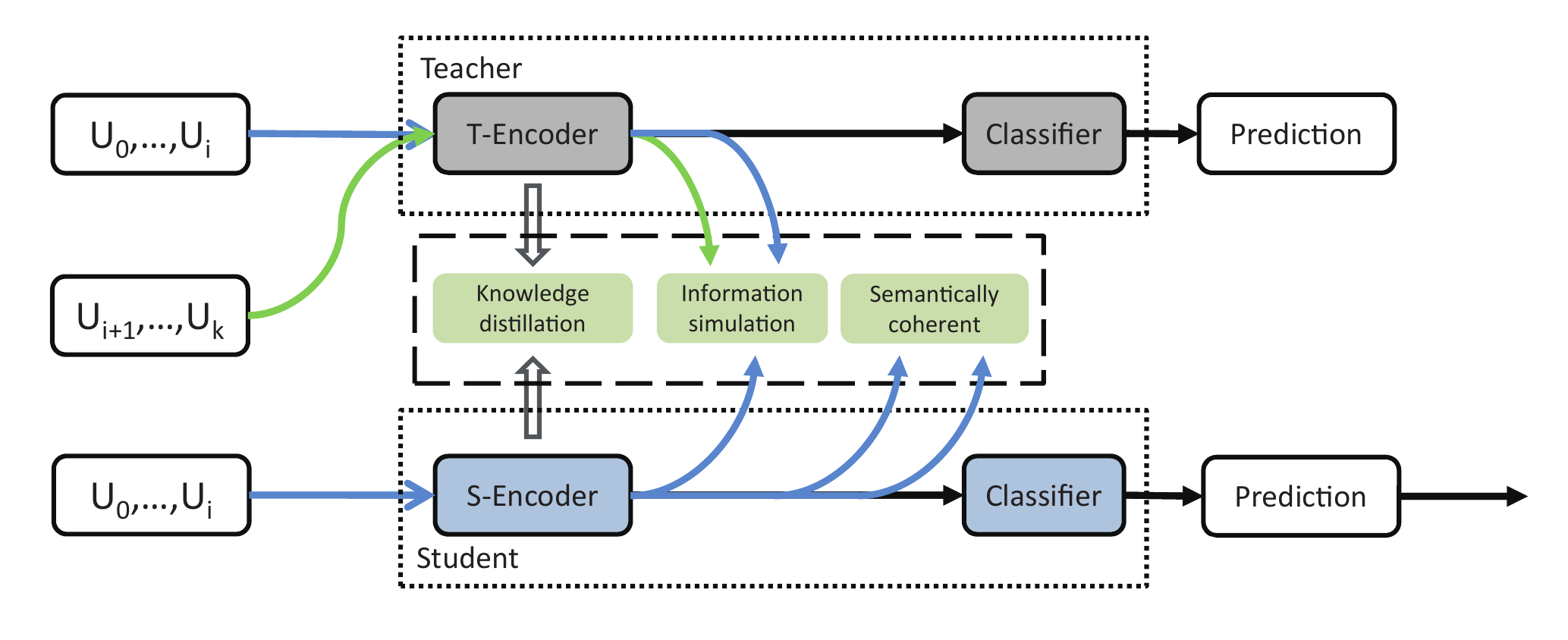}}
\caption{Architecture of the model. The figure contains a teacher encoder, which can obtain the context and the response, and a student encoder, which has only context as input, where $u_{i+1}$ denotes the response in the current dialogue sample and $k$ denotes the length of the dialogue. In addition, the student is prompted to learn from the teacher through knowledge distillation and hierarchical contrastive learning which contains information simulation loss, and semantic coherence-aware loss.}
\label{fig5}
\vspace{-0.4cm}
\end{figure*}
\section{Model}
The framework of our model is shown in Fig.~\ref{fig5}. We propose a teacher-student framework based on Hierarchical Contrastive Learning, which contains two parts: knowledge distillation and hierarchical contrastive learning which consists of two different contrastive learning.

\subsection{Knowledge Distillation}
Existing studies cannot effectively predict topic shifts due to the lack of future information. To address this problem, we introduced a teacher-student framework for dialogue topic shift detection. The student side learns an implicit way of topic prediction from the teacher side through knowledge distillation.

In this framework, we employ a pre-trained model as an encoder on both sides to obtain semantic representations of the dialogues. Besides, we share the encoder weights on the teacher when encoding the context and the response information. For instance, the representation of $[CLS]$ of the last hidden layer state is taken as a dialogue-level representation $H$ as follows.
\begin{equation}
  H^T_{C,i}=E_T\left( X_{C,i}\right)
\end{equation}
\begin{equation}
  H^S_{P,i}=E_S\left( X_{P,i}\right)
\end{equation}
where $H \in R^{N\times d} ,i \in I =\left\{1,2,...,N\right\}$ denotes the index of samples in the batch, $C \in \left\{P,F,A\right\}$ where $P$ represents the context, $F$ represents the response, $A$ represents the full dialog. $E$ represents the encoder, $T$ represents the teacher, and $S$ represents the student. \\
\indent On the teacher side, we connect the dialogue-level representations and then feed them into the linear layer to obtain the final detection results while on the student as follows.\\
\begin{equation}
  Z^T_{i}=W_T\lbrack H_{P,i};H_{F,i};H_{A,i}\rbrack
\end{equation}
\begin{equation}
  Z^S_{i}=W_S H_{P,i}
\end{equation}
where $W_T$ and $W_S$ denote the linear layers on the teacher and student sides, respectively.

In addition to calculating the cross-entropy loss of the final detection results, we establish the mean squared error loss between each hidden layer of the teacher and student encoders.

\subsection{Hierarchical Contrastive Learning}
We consider that the response information learned by knowledge distillation solely is insufficient. And the unbalanced proportion of shift or not in the dialogue corpus makes the model perform poorly in distinguishing different shift cases. Therefore, we propose hierarchical contrast learning, consisting of information simulation loss at the high level and semantic coherence-aware loss at the low level, respectively. Both losses are based on contrast learning, but the former is to strengthen the learning of features and the latter is to alleviate the imbalance of labels.

For information simulation loss, this loss enables active learning for each context representation. And this effectiveness has been demonstrated by several works \cite{dai2021dialogue,oord2018representation,feng2020regularizing}, the incorporation of global information permits local information representation with some predictive information. This helps our encoder obtain a representation with more reliable predictive information.

To alleviate the unbalanced proportion of labels in the dialogues, we propose a semantic coherence-aware loss based on supervised contrast learning (SCL). The main concept of SCL\cite{li2022contrast,gunelsupervised} is to regard the samples of different categories as positive and negative samples from each other to address the issue of significant quantitative imbalance in the dataset\cite{li2017dailydialog}. This loss effectively alleviates the imbalance of shift cases and helps the model further distinguish between different shift cases.

\textbf{High-level Information Simulation Loss}
We build the information simulation loss on both sides so that the context representations can be mapped to the same high-dimensional space. Thus, it is easier to learn the response information to improve final detection results. The following equation can be used to describe this loss.
\begin{equation}
L^M_{ISL}=\sum_{i\in I}\log\frac{exp\left(H^M_{P,i}\cdot H^T_{A,i}\right)}{\sum_{j\in A\left(i\right)}exp\left(H^M_{P,i}\cdot H^T_{A,j}\right)}
\end{equation}
where $A(i)=I-\left\{i\right\}$ denotes the samples in the current batch other than itself, $P$ denotes the context, $A$ denotes the full dialog, and $M \in \left\{T,S\right\}$ where $T$ denotes the teacher, $S$ denotes the student.

\textbf{Low-level Semantic Conherent-aware Loss} For a batch with $N$ training samples, a copy of the dialogue's last hidden state $H$ is made to obtain $\overline{H}$ that is considered as the positive, and its gradient is detached. This results in $2N$ samples, then the semantic coherence-aware loss of all samples in a batch can be expressed as follows.
\begin{equation}
U=\lbrack H;\overline{H}\rbrack
\end{equation}
\begin{equation}
L_{SCL}=\sum_{i\in I}\frac{-1}{\left|P\left(i\right)\right|}\sum_{p\in P\left(i\right)}\log\frac{exp\left(U_i\cdot U_p\right)}{\sum_{a\in A\left(i\right)}exp\left(U_i\cdot U_a\right)}
\end{equation}
where $U\in R^{2N\times d},i\in I=\left\{1,2,...,2N\right\}$ denotes the index of samples in a batch, and $P\left(i\right)=I_{j=i}-\left\{i\right\}$ denotes samples of the same category as $i$ but not itself.
\subsection{Model Training}
We train our model in two steps. The teacher is trained first, and its loss consists of two parts: cross-entropy loss between predictions and manual annotation labels, named $L_{NCE}$. And the information simulation loss for learning response representation is named $L^T_{ISL}$. The overall training loss is as follows.
\begin{equation}
L^T=L_{NCE}+L^T_{ISL}
\end{equation}
\begin{equation}
L^S=L_{NCE}+L_{KD}+L^S_{ISL}+L_{SCL}
\end{equation}

Then, we train the student, which consists of four parts: cross-entropy loss between predictions and manual annotation labels named $L_{NCE}$, knowledge distillation between each hidden layer of both sides named $L_{KD}$, information simulation loss for the dialogue context named $L^S_{ISL}$, and the semantic coherence-aware loss for different shift cases named $L_{SCL}$. The above equation represents its overall training loss. As it proves to be arduous to fine-tune the weights assigned to the four losses, we ultimately opt for equal weighting across all of them.

\section{Experiments and Analysis}
\subsection{Experimental Settings}

Based on the train/validation/test dataset of CNTD we partitioned in Table~\ref{tab2} and previous work on TIAGE \cite{xie2021tiage}, we extract (context, response) pairs from each dialogue as input and the label of response as a target for the response-unknown task. In our experiments, every utterance except the first utterance of the dialogue can be considered as a response. As for evaluation, we report Precision (P), Recall (R), and Micro-F1 scores.

We use BERT as an encoder and fine-tune it during training. For both the TIAGE and CNTD corpus, all pre-trained model parameters are set to default values. We conduct our experiments on NVIDIA GeForce GTX 1080 Ti and NVIDIA GeForce GTX 3090 with batch sizes of 2 and 6 for both CNTD and TIAGE, with the initial learning rates of 2e-5. And we set the epochs of training to 20, and the dropout to 0.5.

For the pre-trained models in the experiment, we apply BERT-base-Chinese and MT5-base to obtain the semantic representation of the dialogues in CNTD, and we apply BERT-base-uncased and T5-base to obtain the semantic representation of the dialogues in TIAGE.
\begin{table}[htbp]
 \setlength\tabcolsep{31pt}
\vspace{-0.2cm}
\caption{Comparison of our model and three baselines on CNTD.}
\begin{center}
\vspace{-0.2cm}
\begin{tabular}{p{1.4cm}ccc}
\hline
Model & \textbf{P}& \textbf{R} & \textbf{F1} \\
\hline
T5 & 27.1 & 46.8 & 34.3 \\
BERT & 55.5 & 43.8 & 48.9 \\
TS & 52.7 & 47.4 & 49.9 \\
Ours & \textbf{56.0} & \textbf{52.0} & \textbf{53.9} \\
\hline
\end{tabular}
\label{tab4}
\end{center}
\vspace{-0.4cm}
\end{table}

\subsection{Experimental  Results}
Dialogue topic shift detection is a new task and there is no complex model available, besides a simple T5 \cite{xie2021tiage} that can be considered as the SOTA model.
Since we employ BERT as our encoder and the T5 model is used in TIAGE, we use the pre-trained models of T5 \cite{raffel2020exploring} and BERT \cite{kenton2019bert} as baselines. 
For BERT, we connect the utterances in the context and separate the last utterance with $[SEP]$. For T5, we also connect utterances in the context and classify the undecidable predicted results to the `not a topic shift' category.

Table~\ref{tab4} shows the performance comparison between our model and the baselines, in which TS denotes our teacher-student model without the hierarchical comparative learning (HCL) and Ours denotes our final model, i.e., the addition of SCL on the student side based on the addition of ISL on both the teacher and student sides.

It can be found that on CNTD, our model achieves a good improvement and improves both precision and recall in comparison with the baselines. Although T5 does not perform poorly on recall, its precision is inadequate in comparison with BERT, and it is clear that T5 is not effective in predicting topics. In contrast, TS improved by 1.0 in Micro-F1 in comparison with BERT, which confirms that the teacher-student framework is effective in introducing response information. As well, Ours improved by 4.0 in micro-F1 in comparison with TS, and also showed significant improvement in P and R, which fully demonstrates that our HCL can improve the model's ability to discriminate between different topic situations.
In particular, our model improves on CNTD by 5.0 in comparison with the best baseline BERT, which shows the effectiveness of our proposed model.

\subsection{Ablation Study}
To verify the effectiveness of the components used in our model, we conduct ablation studies on CTND, and the experimental results are shown in Table \ref{tab12}.

If we remove ISL on the teacher side ($-ISL_S$) or the student side ($-ISL_T$), the performance of the model decreased by 1.5 and 1.3 on the Micro-F1 value, respectively, with the largest decrease after removing the ISL on the student side. Although $-ISL_T$ has the highest precision in predicting topics and lower error probability than $Ours$ and $-ISL_S$. However, it can be seen that adding ISL at both the teacher and student sides can better improve the correct prediction rate. Moreover, if we remove ISL both on the teacher and student side ($-ISL_{TS}$), it achieves a similar performance on Micro-F1, in comparison with $-ISL_{S}$ and $-ISL_{T}$. However, it achieves the highest precision (58.8\%). 

\begin{table}[htbp]
\vspace{-0.1cm}
 \setlength\tabcolsep{32pt}
\caption{Results of our model and its variants on CNTD where ISL, SCL, and HCL refer to the information simulation Loss, semantic conherent-aware Loss, and hierarchical contrastive Learning, respectively, and T, S, and TS refer to teacher side, student side and both of them. }
\begin{center}
\vspace{-0.2cm}
\begin{tabular}{p{1.4cm}ccc}
\hline
& \textbf{P}& \textbf{R} & \textbf{F1} \\
\hline
Ours & 56.0 & 52.0 & \textbf{53.9} \\
$-ISL_S$ & 53.6 & 50.8 & 52.2 \\ 
$-ISL_T$ & 56.1 & 49.3 & 52.6 \\ 
$-ISL_{TS}$ & \textbf{58.8} & 47.0 & 52.3 \\ 
$-SCL$ & 51.6 & \textbf{53.1} & 52.4 \\ 
$-HCL$ & 52.7 & 47.4 & 49.9 \\ 
\hline
\end{tabular}
\label{tab12}
\end{center}
\vspace{-0.2cm}
\end{table}

If we remove SCL (-SCL) or HCL (-HCL) from our model, the Micro-F1 value of the models -SCL and -HCL drop from 53.9 to 52.4 (-1.5) and 49.9 (-4.0), respectively. These results show that our Semantic Conherent-aware Loss(SCL), and Hierarchical Contrastive Learning(HCL) are effective for this task, especially HCL. 

\begin{table}[h]
 \setlength\tabcolsep{27pt}
\caption{Performance on dialogues with the different number of topics.}
\begin{center}
\vspace{-0.2cm}
\begin{tabular}{ccc}
\hline
\textbf{Topic Number} & \textbf{Our Model(F1)}& \textbf{BERT(F1)} \\
\hline
2 & 100 & 66.7\\
3 & 49.0 & 42.6\\
4 & 64.2 & 53.7\\
5 & 58.5 & 46.3\\
6 & 50.3 & 50.7\\
7 & 44.4 & 47.5\\
8 & 44.4 & 40.0\\
9 & 76.9 & 54.5\\
\hline
\end{tabular}
\label{tab5}
\end{center}
\vspace{-0.4cm}
\end{table}

\subsection{Analysis on Different Angles of Performance}
In addition, we explore the performance of the dialogues with different numbers of topics to analyze our model in comparison with BERT, as shown in Table~\ref{tab5}. It can be found that our model has a better performance than BERT on dialogues with fewer topics. Our model gets at least a 6\% improvement in topic shift prediction on dialogues with 2 to 5 topics and obtains above-average performance. And when the number of topics increases to 9, the performance improves because the conversation length is still about 20 and the topics shift more significantly.

In Table~\ref{tab6}, we also investigate the recall of the topic shift detection for various topic turns. Our model is improved for varying degrees across topic turns, with the most significant improvements in turns 7-9. Even in long topic shift cases, our model can obtain an effective boost. However, the performance of our model inevitably decreases compared to short topic shift cases. When there are fewer topic turns, the topic shift situation is simpler, so it is easier to determine. When the length of turns becomes longer and the situation becomes complicated, the topic of long turns has more information so it is easier to identify.

\begin{table}[h]
\vspace{-0.4cm}
 \setlength\tabcolsep{22pt}
\caption{Performances of topic shift with different turn lengths.}
\begin{center}
\begin{tabular}{ccc}
\hline
\textbf{Topic Turns} & \textbf{Our Model(Recall)}& \textbf{BERT(Recall)} \\
\hline
1-3 & 56.6 & 53.8\\
4-6 & 30.3 & 21.4\\
7-9 & 28.8 & 11.9\\
10-12 & 40.0 & 32.0\\
13-17 & 40.0 & 30.0\\
\hline
\end{tabular}
\label{tab6}
\end{center}
\vspace{-0.2cm}
\end{table}

\begin{table}[htbp]
\vspace{-0.2cm}
 \setlength\tabcolsep{30pt}
\caption{Results on TIAGE.}
\begin{center}
\begin{tabular}{p{1.4cm}ccc}
\hline
& \multicolumn{3}{c}{TIAGE}\\
\cline{2-4}
& \textbf{P}& \textbf{R} & \textbf{F1}\\
\hline
T5 & \textbf{34.0} & 17.0 & 22.0\\
BERT & 28.1 & 17.9 & 21.7\\
TS & 26.9 & 20.1 & 22.9\\
Ours & 27.4 & \textbf{28.3} & \textbf{27.8}\\
\hline
\end{tabular}
\label{tab9}
\end{center}
\vspace{-0.2cm}
\end{table}
\subsection{Results on English TIAGE}
As shown in Table~\ref{tab9}, it can be found that our model also achieves a good improvement on English TIAGE. Although our model is not the best on precision, we obtain the best performance on both recall and Micro-F1 values, especially on micro-F1 with a 5.8\% improvement over T5. This proves that our model achieves the best performance both in English and Chinese.

\subsection{Case Study and Error Analysis}
We also conducted a case study. The prediction made by our model, the BERT model on the instance, and the manual labels are shown in Table~\ref{tab7}. Compared with the BERT model, it is obvious that our model can accurately anticipate the change of topic in the instances corresponding to the utterances 
"Yes, that's right.","It is, indeed, should pay attention to it." 
etc., belonging to the question-answering scenario. However, if you respond
"Well, the policy has been implemented in place this time." and "And now we are promoting the development of children's creative and practical skills." etc. belonging to the commenting on the previous context scenario, our model or BERT cannot accurately predict the topic shift in this scenario. This shows that detecting the topic shifts in natural dialogue is still challenging.

\begin{table}[htbp]
 \setlength\tabcolsep{29pt}
\caption{The results of BERT, Ours, and Human of different turns where "1" indicates that a topic shift has occurred and "0" indicates the opposite. We omit the lines with all 0.}
\begin{center}

\begin{tabular}{cccc}
\hline
\textbf{Turns} & \textbf{BERT}& \textbf{Ours}& \textbf{Human} \\
\hline
3 & 0 & 1 & 1\\
5 & 0 & 1 & 1\\
11 & 0 & 1 & 1\\
13 & 1 & 0 & 0\\
14 & 0 & 0 & 1\\
16 & 1 & 0 & 0\\
17 & 0 & 1 & 1\\
18 & 1 & 1 & 0\\
19 & 0 & 1 & 1\\
\hline
\end{tabular}
\label{tab7}
\end{center}
\vspace{-0.2cm}
\end{table}

We further analyze the errors of the prediction produced in our experiments. Specifically, we analyzed the example to explore whether the error in the results of this example is prevalent in other dialogues. From Table~\ref{tab7}, we can find that the wrong predictions at $14^{th}$ and $18^{th}$ turn. We predict 
"The teaching equipment must be updated, right?" as `not a topic shift' and 
"Well, thanks to the government!" as `topic shift'. 

We counted the appearance of many errors, and the errors are mainly divided into two categories. One is for the "Introducing a relevant but different topic" type of utterance. It was predicted that no topic shift occurred due to the lack of information about the future of the conversation. The other is the "commenting on the previous context" category. Since this type of response does not affect the integrity of the previous topic, it is mostly predicted to be a topic shift.

\section{Conclusion} Based on the NaturalConv dataset, we create the CNTD dataset with manual annotations, which fill the gap in the Chinese natural dialogues topic corpus. And we propose a teacher-student model based on hierarchical contrastive learning to solve the lack of response information. We introduced response information through a teacher-student framework and constructed information simulation learning in high-level teachers and students and semantic conherent-aware learning in low-level students. The experiment results demonstrate that our model can perform better in dialogue with few topics. However, detecting the long turns topics or the dialogues with more topics remains a complex problem. Our future work will focus on how to better use response information and news information to detect topic shifts in real-time.
\section*{Acknowledgements}
The authors would like to thank the three anonymous reviewers for their comments on this paper. This research was supported by the National Natural Science Foundation of China (Nos.   62276177, 61836007 and 62006167), and Project Funded by the Priority Academic Program Development of Jiangsu Higher Education Institutions (PAPD).
%
%
%
%

\end{document}